# A Computational Investigation on Denominalization

**Zahra Shekarchi**, **Yang Xu**

Department of Computer Science, University of Toronto
[zahra, yangxu]@cs.toronto.edu

## Abstract

Language has been a dynamic system and word meanings always have been changed over times. Every time a novel concept or sense is introduced, we need to assign it a word to express it. Also, some changes have happened because the result of a change can be more desirable for humans, or cognitively easier to be used by humans. Finding the patterns of these changes is interesting and can reveal some facts about human cognitive evolution. As we have enough resources for studying this problem, it is a good idea to work on the problem through computational modeling, and that can make the work easier and possible to be studied in large scale. In this work, we want to study the nouns which have been used as verbs after some years of their emergence as nouns, and find some commonalities among these nouns. In other words, we are interested in finding what potential requirements are essential for this change.

## 1 Introduction

People can easily create and understand new expressions and statements which they have never heard before and understand their meanings as well. This flexibility of our language is called *innovation* [1]. The context in which the expression is mentioned can help the reader or listener to figure out the intended meaning of the writer or speaker. The rules of composition, in linguistics literature, help people to produce and comprehend these novel expressions. One type of these innovations is that we use a noun as a verb in a sentence. For example, if I say *sauna after exercising,* my meaning can easily be comprehended. This is equivalent to *take a sauna after exercising.* Although this form is more frequent, some people may like to use the former as it is shorter and at the same time conveying the meaning. This type of innovation seems cognitively easier for people. One common explanation for many changes in human language is the *principle of least effort* [2]. Based on that, when people speak, among different ways of expressing their meaning, they prefer the easier one. The listeners also like to minimize their effort in understanding and disambiguating. When the nouns have come to be used as verbs, they are called denominal verbs. There are a lot of examples of this type of innovation in English. This is a zero-conversion of a noun to a verb, and these verbs have different meanings including 'use X as an instrument', 'put something in or on X', 'do something in the manner of X', etc [3]. In [1], Clark and Clark state that there is a convention between the speaker and listener; when the speaker utters a denominal verb in a sentence, (s)he thinks that the listener sees the verb picks out a unique kind of state, event, or process. Then, the speaker would be sure that the meaning will be conveyed based on the context and their mutual knowledge between the speaker and the listener.

Some interesting examples in recent years can be the words like 'google', 'email', or 'skype'. Some common examples are words like 'mop' or 'bike'. One of the important aspects of this change is to know which nouns keep their new verb meaning over time and in which cases this change would not be accepted by the language users. Is it possible to predict which nouns are more probable to be used as verbs in the future? As there are good historical resources for investigating on this problem, in this work, we want to find some commonalities among these nouns through computational methods.

The structure of the paper is that we talk about our hypothesis, then we demonstrate how we model our hypothesis and experiments to evaluate it. Then, the results of the experiments will be discussed. At the end we conclude and propose some lines for future work.

## 2 Hypothesis

We hypothesize that we can define some features of the words which are correlated with the denominalization change. The proposed features are word length, word frequency, and number of word senses. As the time is essential in this study, we consider an interval of years (for example from 1500 to 2000). Thus, for frequency we can define two types; accumulated frequency (frequency of the noun from the beginning of the interval to the year it has been used as verb) and recent frequency (the frequency of the noun at the time before the year it has been used as verb). Also, for the number of word senses we consider the number of senses up to the year it has been used as verb.

We also define two measures for the denominalization change in intended interval; a 'change' parameter (change($w$) for each word $w$) which is a binary value which is 1 for change and 0 for showing no change in the usage of a noun, and 'd' (d($w$) for each word $w$) which shows the time distance between the year a noun emerged and the year it was used as verb for the first time. If there is no such change for a word, then the 'd' parameter will not be computed for that word.

At this point, we can find the correlation between features defined above and these measurements. Through the information of this point we are able to examine the predictability of this denominalization change or how each of those features can play a role in the change.

## 3 Experiments

The first step of the work is to consider a set of nouns and examine whether there is a denominalization change for each noun or not. The dataset we use is the British National Corpus (BNC) [4], including 6000+ lemmatized English words. We extract the nouns which have emerged from year 1500 to 2000. The part of speech tags of these words are extracted from Historical Thesaurus of English (HTE) dictionary [5].

The next step is to compute each feature, introduced in previous section, for the nouns in the target set. For word frequency, we use the frequency reported in Google N-gram corpora [6, 7], for both accumulative and recent frequencies (nouns in Google 1-gram English corpus). For the number of word senses, we use HTE for each word in the set and count their number before the year in which the noun was used as verb for the first time.

In the next step, we can compute the Pearson correlation coefficients for each feature and parameter 'd'. Also, we can calculate the T-test for each feature and parameter 'change'. Based on the results of these, we can make a linear regression model for 'd' and a logistic regression model for 'change'.

## 4 Results

In this part we discuss the results of the experiments. As it is described in previous section, we make a target set of nouns. The number of nouns have emerged and the number of changed nouns from year = 1500, 1600, 1700, 1800, and 1860 to year 2000 are shown in table 1. As we see in this table, we have enough data points if we consider the year 1500. But, as we consider further years in the history, the data gets more scattered. So, we repeat the experiments on data by considering the 'from year' = 1800.

For the recent frequency feature, we use three different intervals; 1, 2, and 3 year(s) before the year we find the change. If we consider only 1 year, we can see the correlation between the features and 'd' in figure 1.

| From year | Number of nouns emerged | Number of nouns have changed |
|---|---|---|
| 1860 | 100 | 18 |
| 1800 | 212 | 31 |
| 1700 | 329 | 54 |
| 1600 | 721 | 133 |
| 1500 | 1357 | 291 |

Table 1. Number of nouns emerged and changed in the interval from each year to 2000

The Pearson correlation coefficients are computed and shown in table 2 and 3. The tables demonstrate that there is no difference among the rows and there are consistent results for the recent frequency feature. More importantly, we can see that the p-value scores for word accumulative frequency and word length are not small enough, and more important features are word recent frequency and word sense number.

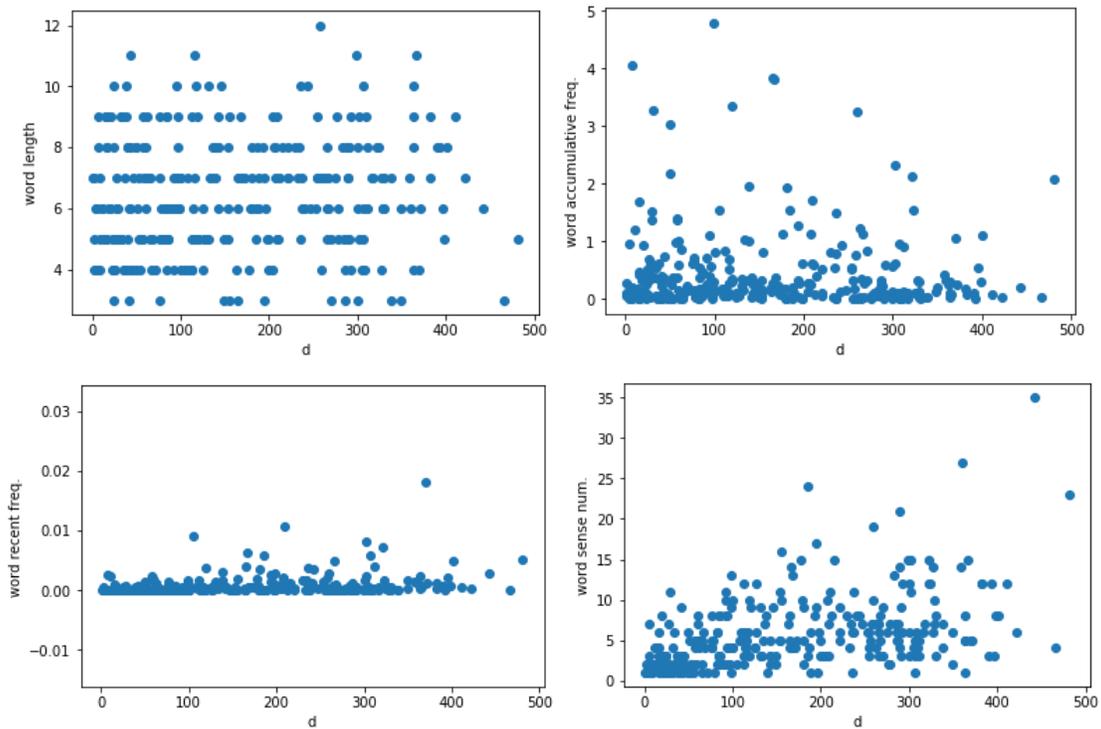

Figure 1. The correlation between word features and 'd'

| Number of years before possible denominalization change | Word length | Word accumulative frequency | Word recent frequency | Word sense number |
|---|---|---|---|---|
| 1 | 0.08,0.18 | -0.04,0.49 | 0.26,8.73e-6 | 0.47,1.53e-17 |
| 2 | 0.08,0.18 | -0.04,0.49 | 0.25,1.21e-5 | 0.47,1.53e-17 |
| 3 | 0.08,0.18 | -0.04,0.49 | 0.25,1.20e-5 | 0.47,1.53e-17 |

Table 2. Pearson correlation coefficients and corresponding p-value between word features and 'd', in interval [1500,2000]

| Number of years before possible denominalization change | Word length | Word accumulative frequency | Word recent frequency | Word sense number |
|---|---|---|---|---|
| 1, 2, 3 | 0.23,0.20 | 0.09,0.63 | 0.51,0.0031 | 0.38,0.035 |

Table 3. Pearson correlation coefficients and corresponding p-value between word features and 'd', in interval [1800,2000]

The mean and standard deviation for each feature is computed and inserted in table 4 and 5. The error plots for each category of nouns which have or have not changed are demonstrated in figure 2 (from year 1500 to 2000). As we can see here, first point is that the variance in the data is too much and this shows high level amount of noise in the data. So, we repeat this experiment for the data in case we consider the interval from 1800 to 2000. The results are in table 6 and we only consider 1 year before the possible change in usage of the word. Although it seems there is less noise in the data and the results are cleaner, we cannot say it is easy to interpret the results. We only can say word length and word recent frequency are more informative and discriminative. Therefore, at this point, we calculate the T-test scores and p-values for each feature and two categories of nouns which have change or do not. This result is shown in table 7. As we see in table 7, word length and word recent frequency have meaningful scores for two types of nouns. This is not almost true for word sense number. For word accumulative frequency we cannot see stable result. If we consider the interval from 1500 to 2000 and consider two and three years before the change in usage, the p-value score for word recent frequency would be 2.74e-6 and 2.31e-6, respectively.

| Number of years before possible denominalization change | Word length | Word accumulative frequency | Word recent frequency | Word sense number |
|---|---|---|---|---|
| 1 | 6.45±1.88 | 0.0044±0.0071 | (8.14±17.55)e-6 | 5.69±4.62 |
| 2 | 6.45±1.88 | 0.0044±0.0071 | (1.64±3.48)e-5 | 5.69±4.62 |
| 3 | 6.45±1.88 | 0.0044±0.0071 | (2.43±5.19)e-5 | 5.69±4.62 |

Table 4. Mean and standard deviation for each word feature and words which have change in usage from year 1500 to 2000

| Number of years before possible denominalization change | Word length | Word accumulative frequency | Word recent frequency | Word sense number |
|---|---|---|---|---|
| 1 | 8.36±2.33 | 0.0032±0.0070 | (1.87±3.68)e-5 | 6.60±5.84 |
| 2 | 8.36±2.33 | 0.0032±0.0070 | (3.74±7.36)e-5 | 6.60±5.84 |
| 3 | 8.36±2.33 | 0.0032±0.0070 | (5.60±11.03)e-5 | 6.60±5.84 |

Table 5. Mean and standard deviation for each word feature and words which do not have change in usage from year 1500 to 2000

| | Word length | Word accumulative frequency | Word recent frequency | Word sense number |
|---|---|---|---|---|
| Words have been used as verb | 6.35±1.94 | 0.0010±0.0016 | (1.91±2.77)e-6 | 3.10±1.92 |
| Words have not been used as verb | 8.60±2.30 | 0.0010±0.0021 | (1.34±1.54)e-5 | 3.33±3.03 |

Table 6. Mean and standard deviation for each word feature and words which do/do not have change in usage from year 1800 to 2000

At the next step, we make a linear regression model for parameter 'd'. The learned weights for each feature is shown in table 8. Based on the Pearson correlation scores, we try omitting some features in the process of training. As we see in this table, word recent frequency has positive weight, which means the more frequent a noun is at a year, the longer it takes to be used as verb in later years. We can see the same pattern for word sense number feature, but this is not true when we consider the interval [1800, 2000]. In table 3, we can see that the p-value for this feature is not small enough.

In the same way, we make a logistic regression model for parameter 'change'. The learned weights for each feature is shown in table 9. Based on T-test p-values from table 7, in different experiments, we do not consider some features and show this by X sign. In this table, we can see a consistent pattern that word length and word recent frequency features have negative weights. It means that the longer a noun is, it is less probable that the noun will be used as verb. Also, the more frequent a noun is, there is less probability that the noun become verb. This is consistent with the result from table 8 for interval [1500, 2000].

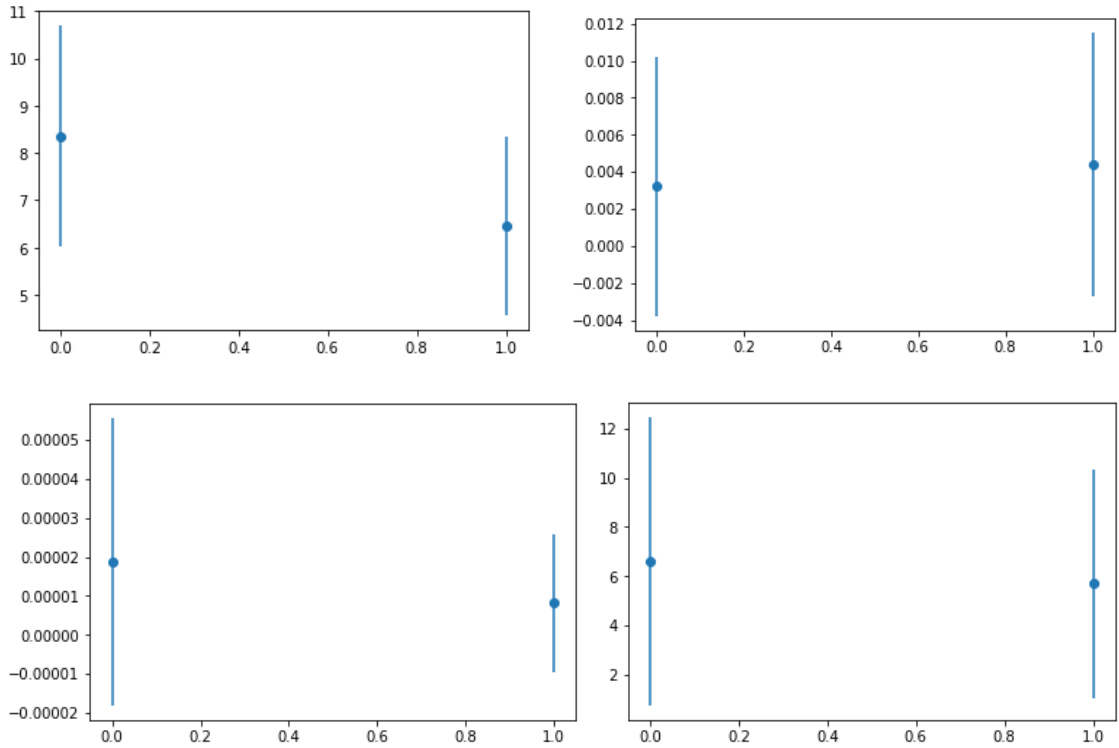

Figure 2. mean and error of each word feature and parameter 'd' (0 for no change and 1 for change), from year 1500 to 2000

|  | Word length | Word accumulative frequency | Word recent frequency | Word sense number |
|---|---|---|---|---|
| T-test p-value From year 1800 | 8.74e-7 | 0.97 | 5.48e-5 | 0.68 |
| T-test p-value From year 1500 | 2.51e-35 | 0.0062 | 2.55e-6 | 0.014 |

Table 7. p-value scores for each word feature, in two intervals of [1500,2000] and [1800,2000]

There is a point about word accumulative frequency; although the Pearson correlation score and T-test p-value for this feature are not significantly meaningful, we can see its weight in linear regression is negative and in logistic regression is positive. This can be interpreted that when the accumulative frequency is high, the word is more accessible to be used as verb over time. Nonetheless, this interpretation cannot be necessarily true, as we see the same sign in each table 8 and 9, we make this discussion about it.

## 6 Conclusion

In this study, we consider some word features and their relation with possible denominalization change. We see that word length has a negative effect on this change. It means that the longer a noun is, the less possible it would be used as verb. But this feature does not have significant relation with the time during which it takes for a noun to become verb. We can conclude that the recent frequency feature is meaningful to show negative correlation with denominalization change and positive correlation with length of change. This means that the more frequent a noun is at a time, the less probable it is to be used as verb in later years.

|  | Word length | Word accumulative frequency | Word recent frequency | Word sense number |
|---|---|---|---|---|
| D (tverb – tnoun) From 1500 | 3.69 | -33.22 | 33.97 | 53.47 |
| D (tverb – tnoun) From 1500 | X | -33.25 | 33.98 | 53.83 |
| D (tverb – tnoun) From 1500 | X | X | 17.18 | 51.60 |
| D (tverb – tnoun) From 1800 | 9.91 | -25.45 | 27.66 | 13.38 |
| D (tverb – tnoun) From 1800 | X | X | 20.07 | -0.66 |

Table 8. learned weights for each word feature in linear regression model. X shows corresponding feature is not considered in training

|  | Word length | Word accumulative frequency | Word recent frequency | Word sense number |
|---|---|---|---|---|
| Change (0/1) From 1500 | -0.84 | 0.77 | -1.39 | 0.017 |
| Change (0/1) From 1500 | -0.83 | 0.78 | -1.39 | X |
| Change (0/1) From 1500 | -0.92 | 0.09 | X | X |
| Change (0/1) From 1500 | -0.91 | X | -0.65 | X |
| Change (0/1) From 1800 | -0.89 | X | -1.87 | X |

Table 9. learned weights for each word feature in logistic regression model. X shows corresponding feature is not considered in training

There are some other experiments that can be considered in this study and make these results clearer. The most interesting one is to consider some categories of nouns which are more probable to have this change in their usage. For instance, instruments, technology products and applications, places, and duration. For each of these categories we can name some nouns which are used as verb; mop, email, sauna, and summer (respectively). We can observe the relation of each word feature with the nouns in these categories and also other categories which maintain their noun usages like animals. These categories can be compared through their syntactic forms and meanings [3]. Furthermore, we can use the Google N-gram corpora instead of HTE and see whether we see the same patterns.

This denominalization change for nouns helps us to produce shorter sentences which can be employed in text summarization. This can be true for the task of paraphrasing, as in this way we say a different sentence with the same meaning. Meanwhile, it is important to see which of these changes are stable and acceptable by language users. Another point which is worth mentioning here is that sometimes these denominal verbs are more attractive to people and this advantage can be considered in advertisements and newspaper titles, where we want to catch people's attention easily, fast, and by minimum number of words.